\documentclass{article}

     \PassOptionsToPackage{numbers, compress, square, comma}{natbib}

\usepackage[final]{ml4mols_2020}

\usepackage[utf8]{inputenc} 
\usepackage[T1]{fontenc}    
\usepackage{hyperref}       
\usepackage{url}            
\usepackage{booktabs}       
\usepackage{amsfonts}       
\usepackage{nicefrac}       
\usepackage{microtype}      
\usepackage{pifont}
\usepackage{enumitem}
\usepackage{cite}

\usepackage{adjustbox}
\usepackage{siunitx}

\usepackage{subcaption}
\usepackage{caption}
\usepackage{ragged2e}
\usepackage{array,multirow}

\usepackage{booktabs}
\usepackage[table]{xcolor}

\definecolor{color-A65B73}{HTML}{A65B73}
\definecolor{color-EB9706}{HTML}{EB9706}
\definecolor{color-4A9995}{HTML}{4A9995}
\definecolor{color-204E61}{HTML}{204E61}

\usepackage{graphicx}
\usepackage{cite}
\newcommand{\molbert}{\textsc{MolBert}}

\usepackage[abbreviations]{glossaries-extra}
\glssetcategoryattribute{acronym}{indexonlyfirst}{true}
\glsdisablehyper
\newabbreviation{qsar}{QSAR}{Quantitative Structure-Activity-Relationship}
\newabbreviation{nlp}{NLP}{Natural Language Processing}
\newabbreviation{ecfp}{ECFP}{Extended-Connectivity Fingerprint}
\newabbreviation{bert}{BERT}{Bidirectional Encoder Representations from Transformers}
\newabbreviation{smiles}{SMILES}{Simplified Molecular Input Line Entry System}
\newabbreviation{svm}{SVM}{Support Vector Machine}

\makeatletter
\def\fixedlabel#1#2{%
  \@bsphack%
  \protected@write\@auxout{}%
         {\string\newlabel{#1}{{#2}{\thepage}}}%
  \@esphack}
\makeatother

\title{Molecular representation learning with language models and domain-relevant auxiliary tasks}

\author{ 
  Benedek Fabian
  \And
  Thomas Edlich
  \And
  H\'el\'ena Gaspar
  \And
  Marwin Segler
  \AND 
  Joshua Meyers
  \And
  Marco Fiscato
  \And
  Mohamed Ahmed
  \AND
  BenevolentAI\\
  4-8 Maple St, Bloomsbury, London W1T 5HD\\
  \texttt{<firstname.lastname>@benevolent.ai}
}

\begin{document}

\maketitle

\begin{abstract}
We apply a Transformer architecture, specifically BERT, to learn flexible and high quality molecular representations for drug discovery problems, and study the impact of using different combinations of self-supervised tasks for pre-training. Our results on established Virtual Screening and QSAR benchmarks show that: i) The selection of appropriate self-supervised task(s) for pre-training has a significant impact on performance in subsequent downstream tasks such as Virtual Screening. ii) Using auxiliary tasks with more domain relevance for Chemistry, such as learning to predict calculated molecular properties, increases the fidelity of our learnt representations. iii) Finally, we show that molecular representations learnt by our model `\molbert' improve upon the current state of the art on the benchmark datasets.

\end{abstract}

\section{Introduction}
\vspace{-0.3cm}

Molecular representations underpin predictive, generative and analytical tasks in drug discovery~\citep{david2020molecular}. The choice of a suitable representation can drastically impact the efficiency of discovering a novel drug candidate. For instance, applications such as \textit{Virtual Screening} and \textit{\gls{qsar}} modeling rely on the availability of effective molecular representations~\citep{vamathevan2019applications}.

Language models have been applied to text-based molecular representations such as \gls{smiles}~\citep{weininger1988smiles}. They show impressive performance across a range of domain applications including molecular property~\citep{jastrzkebski2016learning,winter2019learning} and reaction prediction problems~\citep{liu2017retrosynthetic,schwaller2019molecular}, as well as generative tasks~\citep{segler2018generating}. Numerous strategies have been explored to encourage learning of high quality representations with language models including input reconstruction~\citep{fourches2019inductive, maziarka2020molecule, smilesbert2019wang}, whereby a model learns to predict masked or corrupted tokens; and input translation~\citep{winter2019learning, morris2020moltransformembed}, where the goal is to translate the input to another modality or representation. Further improvements have been made by incorporating calculated molecular properties into the representation, either by concatenating with the learnt representations \citep{dmpnn2019yang}, or through devising pre-training schemes~\citep{chemnet2018goh}. Finally, a range of model architectures have been explored including  autoencoders~\citep{winter2019learning}, RNNs~\citep{fourches2019inductive} and transformers~\citep{maziarka2020molecule,smilesbert2019wang}.

Aside from the modal limitations of representing molecules as strings, a drawback to learning from text-based molecular representations is introduced by the ambiguity of linearizing the molecular graph~\citep{randic-isomorphism:77}. In the case of \gls{smiles}, many valid sequences may represent the same molecule depending on the traversal path of the molecular graph. This ambiguity has led to the development of canonicalization algorithms~\citep{weisfeiler-lehman:68, schneider2015canonicalize} which, while practical, introduce artifacts to linearized \gls{smiles} such that a language model may be distracted by the rules of canonicalization. Previous works have shown the benefits of learning using permutations of \gls{smiles}~\citep{Bjerrum:2018ew}.

In this work, we evaluate the application of the widely used \gls{bert}~\citep{bert2018devlin} architecture for the generation of molecular representations. We explore the impact of employing a range of domain-relevant auxiliary tasks during pre-training and evaluate the produced learnt representations on downstream Virtual Screening and \gls{qsar} benchmarks. Code and pre-trained models are available at \href{https://github.com/BenevolentAI/MolBERT}{https://github.com/BenevolentAI/MolBERT}.

\section{\molbert}\label{methods}
\begin{figure}[t]
    \centering
    \includegraphics[width=0.68\textwidth]{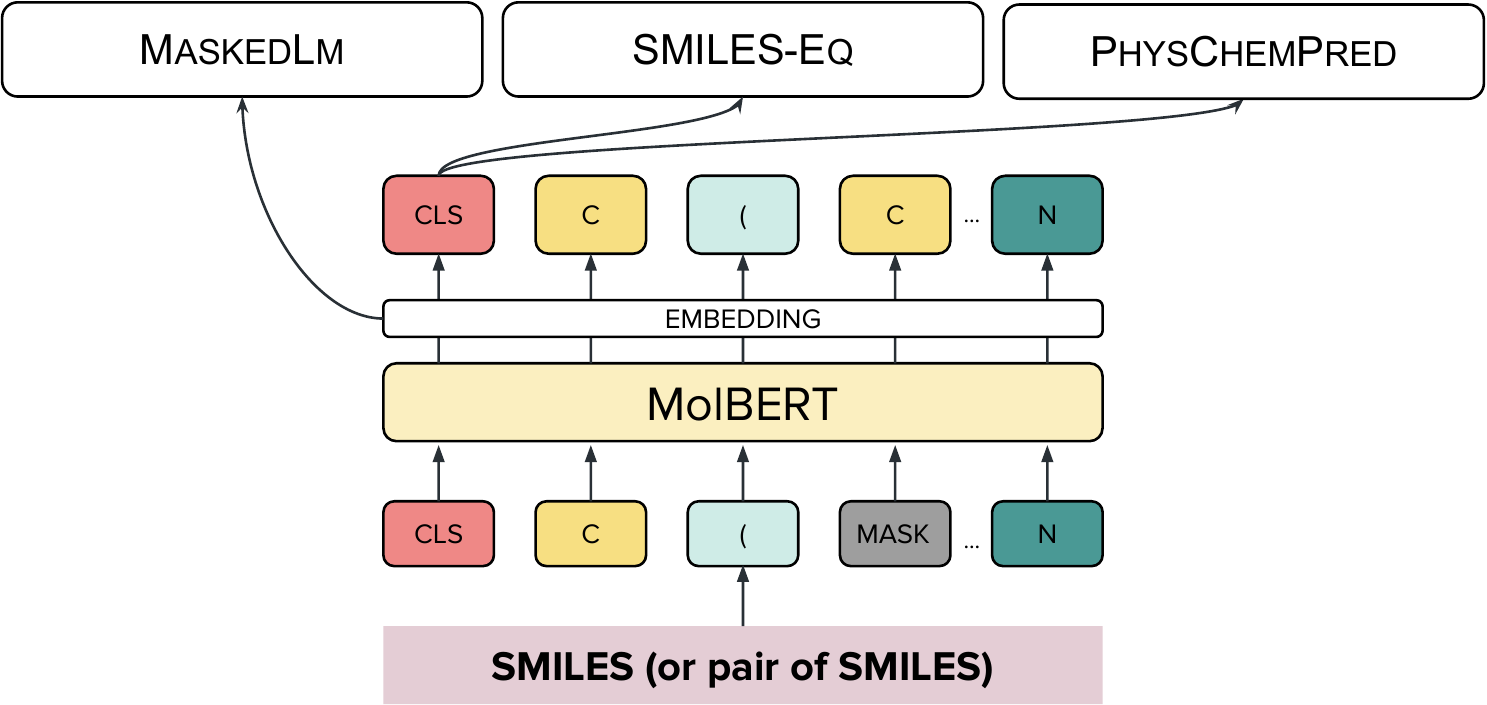}
    \caption{\label{fig:molbert} Diagram of \molbert~illustrating the various auxiliary tasks utilized for pre-training.}
    \vspace{-0.4cm}
\end{figure}

\vspace{-0.3cm}
\molbert, as depicted in~\autoref{fig:molbert}, is a bidirectional language model that uses the \gls{bert} architecture~\citep{bert2018devlin}. To understand the impact of pre-training with different domain-relevant tasks on downstream applications, we experiment with the following set of self-supervised tasks:
\vspace{-0.2cm}
\begin{description}
    [leftmargin=0.3cm,
    style=unboxed,
    labelsep=-0.05cm,
    itemsep=-0.05cm,
    align=left,  font=\bf]
    \item [Masked language modeling (\textsc{MaskedLM}):~] ~The canonical task proposed by BERT, whereby the model is trained to predict the true identity of masked tokens. The task is optimized using the cross-entropy loss between the sequence output and the masked tokens of the input.
    \item [SMILES equivalence (\textsc{SMILES-Eq}):~] ~Given an input of two SMILES where the second is, with equal probability, either randomly sampled from the training set or a synonymous permutation of the first SMILES, the task is to predict whether the two inputs represent the same molecule. This is a binary classification task, which may be traditionally solved by comparing the canonicalized molecular graphs of the inputs. It is optimized using the cross-entropy loss.
    \item [Calculated molecular descriptor prediction
    (\textsc{PhysChemPred}):~] ~Using RDKit~\citep{LandrumRDKit2020_03_2} we are able to calculate a simple set of real-valued descriptors of chemical characteristics and composite models of physicochemical properties for each molecule (see~\autoref{app:physchem}). The goal of this task is to predict the normalized~\citep{yang2019analyzing} set of descriptors for each molecule. The task is optimized using the mean squared error over all predicted values.
\end{description}
\vspace{-0.2cm}
The final loss is given by the arithmetic mean of all individual task losses.
 
{\bf Fine-tuning}\\ 
The representations learnt after pre-training \molbert~may be applied to downstream tasks in several manners. Here we explore: i) using representations directly for similarity search, ii) training a downstream model; in this case, following~\citet{winter2019learning}, we use a \gls{svm}, and finally iii), following~\citep{bert2018devlin}, we add an explicit downstream task head in the form of a newly initialized network which is then optimized.

{\bf Evaluation methodology}\label{methods:evaluation}\\
We evaluate \molbert~on two downstream applications: \textit{Virtual Screening} and \textit{\gls{qsar}}. In Virtual Screening 
we are typically interested in selecting from a pool of candidate compounds the ones that best satisfy a property of interest, such as the likelihood of binding to a particular drug target. Alternatively, in \gls{qsar} we are interested in learning to predict a given molecular property, such as the binding affinity to a target of interest. We evaluate \molbert~when used directly and when specialized for downstream applications using two established benchmarks:
\vspace{-0.15cm}
\begin{description}
    [leftmargin=0.3cm,
    style=unboxed,
    labelsep=-0.05cm,
    itemsep=-0.05cm,
    align=left,  font=\bf]
    \item [{\emph{Virtual Screening:}}~] we use the filtered virtual screening subset (version 1.2) from the RDKit benchmarking platform~\citep{riniker2013open, riniker2013heterogeneous}. The benchmark is made up of $69$ datasets, each corresponding to an individual protein target. Each dataset consists of a small number of active molecules amongst a larger number of target specific decoys. The benchmark measures how well-suited molecular representations are to retrieving active compounds from the pool, given a fixed number of query molecules ($n=5$)~-- for details see~\citet{riniker2013open}.

    Retrieval is performed according to the cosine distance between \molbert~embeddings. Following~\citet{riniker2013open}, we report results using: i) the Area Under the Curve of the Receiver Operating Characteristic (AUROC) and ii) the Boltzmann-Enhanced Discrimination of ROC (BEDROC, $\alpha=20$) -- a widely used early enrichment metric that assigns higher weight to the first $\alpha\%$ of molecules retrieved from the pool according to the Boltzmann distribution~\citep{truchon2007evaluating}.
    
    \item [{\emph{\gls{qsar}:}}~] we use a subset of the MoleculeNet benchmark suite~\citep{wu2018moleculenet}, which consists of datasets of varying size for a range of \gls{qsar} problems. Specifically we report regression results for the ESOL, FreeSolv and Lipophilicity datasets, as well as classification results for the BACE, BBBP and HIV datasets. Since MoleculeNet does not provide explicit training, validation and test folds for all datasets, we use folds provided by ChemBench~\citep{charleshen_2020_4054866} to present reproducible results.
\end{description}



\section{Experimental Evaluation}\label{results}
\vspace{-0.3cm}
In this section, we first present an ablation on \molbert~using the different pre-training tasks using the Virtual Screening datasets, then report our results in the Virtual Screening and \gls{qsar} benchmarks.

\begin{table}[t]
  \begin{minipage}[c]{\columnwidth}
      \footnotesize
      \centering
      \resizebox{\columnwidth}{!}{
      \begin{tabular}{ccc|cccc}
            \toprule
             & & & \multicolumn{2}{c}{w/ permutation}  & \multicolumn{2}{c}{w/o permutation} \\
            \midrule
            MaskedLM &            PhysChemPred &           SMILES-Eq & AUROC  & BEDROC20  & AUROC  & BEDROC20 \\
            \midrule
            \ding{51}     &  \ding{51} &  \ding{51} &  $0.685 \pm 0.069$ &  $0.246 \pm 0.041$ &  $0.707 \pm 0.059$ &  $0.280 \pm 0.042$\\
            \ding{51}     &  \ding{51} &  \ding{55} &  $\mathbf{0.738 \pm 0.060}$ &  $\mathbf{0.323 \pm 0.071}$ &  $\mathbf{0.740 \pm 0.066}$ &  $\mathbf{0.322 \pm 0.065}$ \\
            \ding{51}     &  \ding{55} &  \ding{51} &  $0.483 \pm 0.092$ &  $0.092 \pm 0.069$ &  $0.493 \pm 0.068$ &  $0.108 \pm 0.070$ \\
            \ding{55}     &  \ding{51} &  \ding{51} &  $0.476 \pm 0.077$ &  $0.064 \pm 0.034$ &  $0.514 \pm 0.165$ &  $0.084 \pm 0.014$ \\
            \ding{51}     &  \ding{55} &  \ding{55} &  $0.696 \pm 0.058$ &  $0.283 \pm 0.077$ &  $0.676 \pm 0.060$ &  $0.250 \pm 0.073$ \\
            \ding{55}     &  \ding{51} &  \ding{55} &  $0.719 \pm 0.057$ &  $0.293 \pm 0.071$ &  $0.716 \pm 0.061$ &  $0.290 \pm 0.076$ \\
            \ding{55}     &  \ding{55} &  \ding{51} &  $0.129 \pm 0.067$ &  $0.005 \pm 0.037$ &  $0.508 \pm 0.068$ &  $0.048 \pm 0.035$ \\
            \bottomrule
      \end{tabular}
      }
  \caption{The impact of pre-training \molbert~with different auxiliary task combinations on the Virtual Screening benchmark performance. Best values are in bold, higher values are better.}
  \label{rdkit-ablation}
  \end{minipage}
  \hfill
  \begin{minipage}[c]{\columnwidth}
    \centering
    \small
    \resizebox{0.55\columnwidth}{!}{
    \begin{tabular}{l|cc}
        \toprule
         & AUROC & BEDROC20 \\
        \midrule
        \molbert (100 epochs)     &  $\mathbf{0.743 \pm 0.062}$ &   $\mathbf{0.344 \pm 0.062}$\\
        CDDD     &  $0.725 \pm 0.057$ &   $0.310 \pm 0.080$\\
        RDKit descriptors     &  $0.633 \pm 0.027$ &   $0.217 \pm 0.000$\\
        ECFC4     &  $0.603 \pm 0.056$ &   $0.170 \pm 0.079$\\
        \bottomrule
    \end{tabular}
    }
    \caption{Results for Virtual Screening using the RDKit benchmarking platform.}
    \label{rdkit-overall}
  \end{minipage}
  \vspace{-0.8cm}
\end{table}


\textbf{Pre-training dataset:} All models are pre-trained using the GuacaMol benchmark dataset~\citep{guacamol} consisting of $\sim$1.6M compounds curated from ChEMBL~\citep{gaulton2017chembl}. We use the training and validation splits which consist of 80\% and 5\% of the data, respectively. 

\textbf{Models:} \molbert~models are implemented using the Hugging Face transformers library~\citep{wolf2020huggingfaces}, and use the BERT-Base architecture (12 attention heads, 12 layers, 768 dimensional hidden layer, $\sim$85M parameters). We use the Adam optimizer~\citep{kingma2014adam} with a learning rate of \num{3e-5}, and train for 20 epochs, except for the final model, which we train for 100 epochs. The average pre-training time for \molbert ~was $\sim$40 hours using 2 GPUs and 16 CPUs.

\textbf{Fine-tuning:} All experiments use the same base model, the best model from the ablation trained for 100 epochs. To ease reproducibility we use the~\gls{svm} parameters published in~\citet{winter2019learning} ($C=5.0$, RBF kernel) and all fine-tuning networks are a single linear layer attached to the pooled output. To generate molecular embeddings we use the pooled output, except where \textsc{MaskedLM} is the only task used for pre-training, in which case we use the average of the sequence output, since the pooled output has no dependent tasks.

\molbert~is trained using a fixed vocabulary of $42$ tokens and a sequence length of $128$ characters. In line with~\citep{bert2018devlin}, all tasks are trained by masking 15\% of the tokenized input. To support the use of arbitrary length input SMILES at inference time, we use relative positional embeddings as described by~\citet{transformerxl}.

\begin{table}[t]
  \centering
  \footnotesize
  \resizebox{0.9\columnwidth}{!}{
    \begin{tabular}{l|ccccc}
        \multicolumn{6}{c}{Regression Datasets: RMSE} \\
        \toprule
         &      RDKit (norm) &             ECFC4 &              CDDD &           \molbert & \molbert~(finetune) \\
        \midrule
       ESOL     &  $0.687 \pm 0.08$ &  $0.902 \pm 0.06$ &  $0.567 \pm 0.06$ &  $0.552 \pm 0.07$ &   $\mathbf{0.531 \pm 0.04}$ \\
FreeSolv &  $1.671 \pm 0.45$ &  $2.876 \pm 0.38$ &  $1.456 \pm 0.43$ &  $1.523 \pm 0.66$ &   $\mathbf{0.948 \pm 0.33}$ \\
Lipop    &  $0.738 \pm 0.04$ &  $0.770 \pm 0.03$ &  $0.669 \pm 0.02$ &  $0.602 \pm 0.01$ &   $\mathbf{0.561 \pm 0.03}$ \\
        \bottomrule
  
    \multicolumn{6}{c}{Classification Dataset: AUROC} \\
    \toprule
 &      RDKit (norm) &             ECFC4 &              CDDD &           MolBERT & MolBERT (finetune) \\
\midrule
BACE    &  $0.831 \pm 0.00$ &  $0.845 \pm 0.00$ &  $0.833 \pm 0.00$ &  $0.849 \pm 0.00$ &   $\mathbf{0.866 \pm 0.00}$ \\
BBBP    &  $0.696 \pm 0.00$ &  $0.678 \pm 0.00$ &  $0.761 \pm 0.00$ &  $0.750 \pm 0.00$ &   $\mathbf{0.762 \pm 0.00}$ \\
HIV     &  $0.708 \pm 0.00$ &  $0.714 \pm 0.00$ &  $0.753 \pm 0.00$ &  $0.747 \pm 0.00$ &   $\mathbf{0.783 \pm 0.00}$ \\
\bottomrule
  \end{tabular}}
  \vspace{0.15cm}
  \caption{QSAR results on regression and classification tasks from MoleculeNet. Best values are in bold. $\pm$ indicates standard deviation over cross-validation splits. First four columns are generated using the SVM, while the last column refers to fine-tuning a new task head.}
  \label{tab:moleculenet-results}
  \vspace{-0.8cm}
\end{table}

\subsection*{Results}\label{sec:results}
\vspace{-0.2cm}
{\bf Ablation study:} We first analyze the utility of using different combinations of tasks for pre-training. From \autoref{rdkit-ablation}, the main observations are that: i) The \textsc{PhysChemPred} task has the highest impact on the performance metrics, with an average \textsc{BEDROC20} of $0.292$ when using the \textsc{PhysChemPred} task alone (with and without permutations)  \textit{versus}  $0.266$ for the \textsc{MaskedLM} alone. ii) Although the best performing model is trained on both the \textsc{PhysChemPred} and \textsc{MaskedLM}, the additive gain from the \textsc{MaskedLM} task is relatively minor; $+0.031$ on average for the \textsc{BEDROC20} metric. iii) The addition of the \textsc{SMILES-Eq} task slightly but consistently decreases performance.



Given the effectiveness of pre-training with the~\textsc{PhysChemPred} task, we explored the impact of grouping the $200$ calculated descriptors into disjoint related subsets. \autoref{subset-ablation} shows that, although using \textsc{All} the descriptors achieves the best overall result, the \textsc{surface} properties of a molecule provide a very competitive supervision task using only 25\% of the descriptors.


{\bf Virtual Screening:} \autoref{rdkit-overall} compares the performance of \molbert~when trained for 100 epochs with the best performing auxiliary task combination,  with three baseline methods. i)~\textsc{CDDD}~\citep{winter2019learning}, a neural model that achieves the current state-of-the-art results for this benchmark. ii)~The RDKit calculated physicochemical descriptors~\citep{LandrumRDKit2020_03_2} used for the \textsc{PhysChemPred} task during pre-training. iii)~Extended Connectivity Fingerprints with a diamater of 4 (\textsc{ECFC4}), one of the most commonly used descriptors in drug discovery. Our results show that \molbert~outperforms all other descriptors both in terms of overall classification and early enrichment. A detailed breakdown of \textit{per}-dataset results is given in~\autoref{fig:rdkit-lineplot-auroc} for the BEDROC20 and  AUROC, respectively. Finally, upon closer investigation as to the benefit of input permutation and calculated molecular property prediction, we observed that these strategies enable \molbert~to organize the learnt embeddings. More concretely, pre-training with the \textsc{PhysChemPred} task and input permutation leads to models which on average assign a lower pairwise similarity to non-identical compounds; see~\autoref{sec:consistency}.

{\bf \gls{qsar}:} To further evaluate the usefulness of \molbert~embeddings, we build \gls{qsar} models for datasets from MoleculeNet~\citep{wu2018moleculenet} and include models built using CDDD~\citep{winter2019learning}, ECFC4~\citep{rogers2010extended} and the normalized RDKit calculated physicochemical descriptors~\citep{LandrumRDKit2020_03_2} as baselines. As described in Section~\ref{methods:evaluation}, we train an \gls{svm} (implemented using \texttt{sklearn}~\citep{scikit-learn}) using each molecular representation, and compare against fine-tuning \molbert.

From the results in~\autoref{tab:moleculenet-results}, we see that neural models substantially outperform traditional molecular representations (RDKIT and ECFC4). Furthermore, finetuned \molbert~models achieve the best performance in all of the six benchmark datasets. We also observe that \molbert~representations combined with a SVM outperform the other descriptors on three of the six tasks.


\section{Conclusions}
\vspace{-0.3cm}

We have introduced \molbert, a language model for learning molecular embeddings using \gls{bert}. We investigated the impact of pre-training with domain-relevant auxiliary tasks and found that the choice of self supervision task significantly impacts the performance on downstream tasks. Nevertheless, with the right set of tasks \molbert~ achieves state-of-the-art performance on established \textit{Virtual Screening} and \textit{\gls{qsar}} benchmarks.
We leave to future work the exploration of how to use \molbert~for learning representations of other entities such as proteins~\citep{simonovsky, Alley:2019fe, Kim2020deeppcm}, along with further developments in our learning strategies~\citep{pretraingraphs2019pande}.

\medskip
    
\small
\newpage
\bibliography{main}

\newpage
\appendix
\section*{Appendix}
\renewcommand\thefigure{A\arabic{figure}} 
\renewcommand\thetable{A\arabic{table}} 
\setcounter{figure}{0}
\setcounter{table}{0}

\section{Virtual Screening benchmark: performance \textit{per} target}

\begin{figure}[ht!]
    \includegraphics[width=\linewidth,clip]{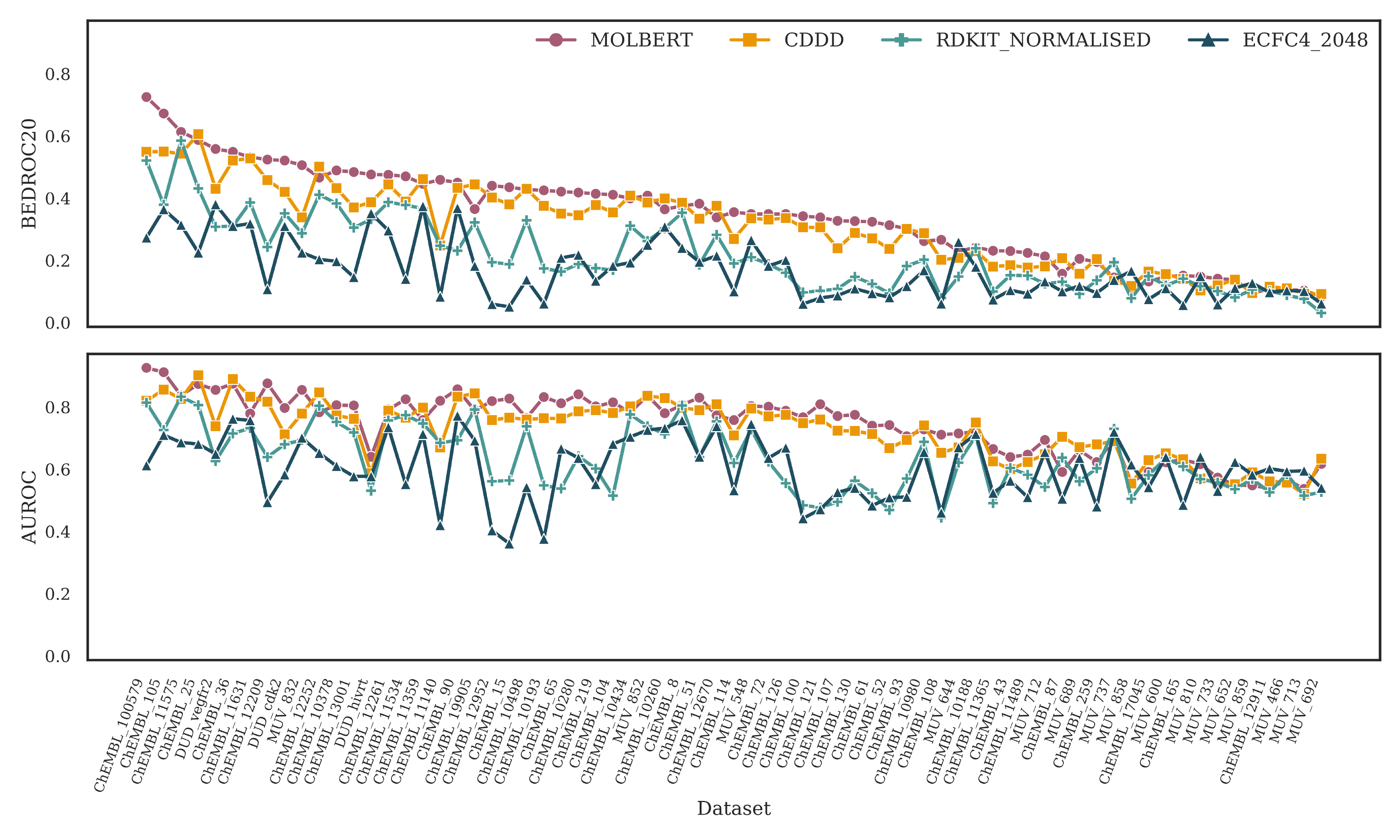} 
    \caption{BEDROC20 and AUROC performance for each of the target datasets in the Virtual Screening benchmark~\citep{riniker2013open} sorted based on the BEDROC20 metric. We report results for the best performing \molbert~model in our ablation study (see~\autoref{rdkit-ablation}) trained for 100 epochs.}\label{fig:rdkit-lineplot-auroc} 
\end{figure}

\autoref{fig:rdkit-lineplot-auroc} shows results for \molbert~and other baseline methods on the individual Virtual Screening datasets listed in~\citep{riniker2013open}. The datasets are sorted based on the  \textsc{BEDROC20} enrichment metric. We observe that \molbert~displays superior performance for 45 of 69 targets and performs competitively in all other cases.

\section{Auxiliary task induced biases in \molbert~models}\label{sec:consistency}
\begin{table}[t]
\begin{tabular}{ c>{\tiny \texttt}p{2in}|>{\tiny \texttt}p{3.in}}
\toprule
  \multicolumn{2}{c}{\small \textbf{Seed Molecule}} & \small \textbf{Permutations} \\
\midrule
                                                        &  & \\
 \multirow{2}{*}{ \cellcolor{color-4A9995!95} \rotatebox[origin=c]{90} {Palbociclib~\quad}} & 

  \begin{flushleft}
  \raisebox{-0.45\height}{\includegraphics[trim={-1.5cm 0 0 0},clip,scale=0.5]{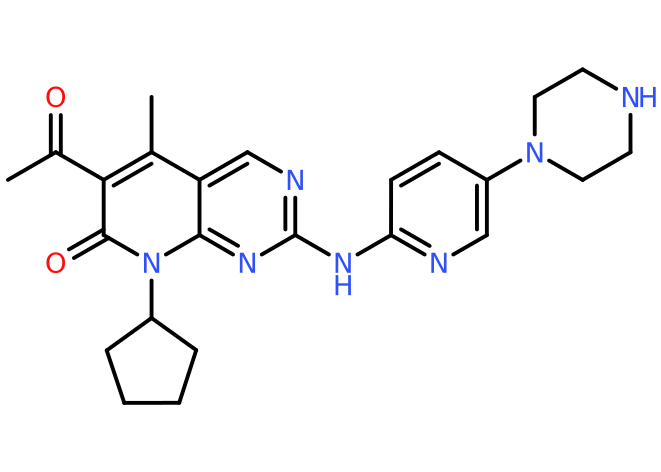}} \\
     C1(NC2=NC=C(N3CCNCC3)C=C2)=NC=C2C(C)=C(C(C)\-=O)C(=O)N(C3CCCC3)\-C2=N1   
    \end{flushleft}  & 
    \begin{flushleft}
     C1(NC2=NC=C(N3CCNCC3)C=C2)=NC=C2C(C)=C(C(C)=O)C(=O)N(C3CCCC3)C2=N1\\
     C1CCC(N2C3=NC(NC4=NC=C(N5CCNCC5)C=C4)=NC=C3C(C)=C(C(=O)C)C2=O)C1\\
     C1(C)=C(C(C)=O)C(=O)N(C2CCCC2)C2=NC(NC3=NC=C(N4CCNCC4)C=C3)=NC=C12\\
     C1C(N2C3=NC(NC4=NC=C(N5CCNCC5)C=C4)=NC=C3C(C)=C(C(=O)C)C2=O)CCC1\\
     N1=C(NC2=NC=C3C(C)=C(C(=O)C)C(=O)N(C4CCCC4)C3=N2)C=CC(N2CCNCC2)=C1\\
     C1=CC(N2CCNCC2)=CN=C1NC1=NC=C2C(C)=C(C(=O)C)C(=O)N(C3CCCC3)C2=N1\\
     C1NCCN(C2=CN=C(NC3=NC=C4C(C)=C(C(C)=O)C(=O)N(C5CCCC5)C4=N3)C=C2)C1\\
     C1CCC(N2C3=NC(NC4=NC=C(N5CCNCC5)C=C4)=NC=C3C(C)=C(C(C)=O)C2=O)C1\\
     C1CN(C2=CN=C(NC3=NC=C4C(=N3)N(C3CCCC3)C(=O)C(C(=O)C)=C4C)C=C2)CCN1\\
     N1(C2=CN=C(NC3=NC=C4C(=N3)N(C3CCCC3)C(=O)C(C(C)=O)=C4C)C=C2)CCNCC1\\
    \end{flushleft} \\
                                                       &  & \\
                                                       \hline
                                                       & & \\
 \multirow{2}{*}{ \cellcolor{color-A65B73!95} \rotatebox[origin=c]{90} {Seliciclib~\quad}} & 
   \begin{flushleft}
  \raisebox{-0.45\height}{\includegraphics[trim={-1cm 0 0 0},clip,scale=0.5]{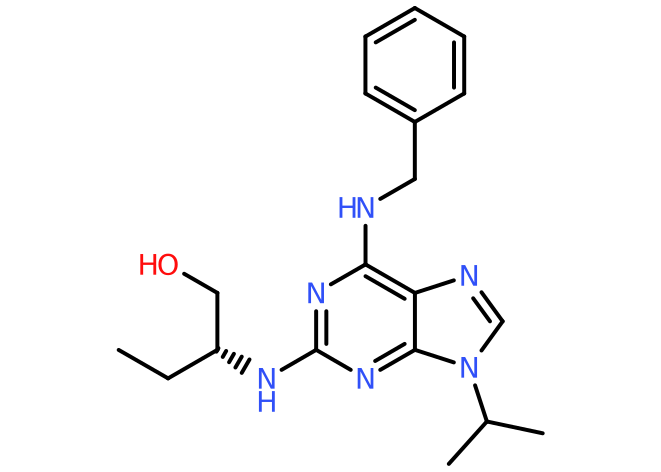}} \\
     CC[C@H](CO)NC1=NC(NCC2=CC=CC=C2)=C2C(=N1)\-N(C(C)C)C=N2  
    \end{flushleft}  & 
    \begin{flushleft}
    C(NC1=C2C(=NC(N[C@H](CC)CO)=N1)N(C(C)C)C=N2)C1=CC=CC=C1\\
    C(NC1=C2N=CN(C(C)C)C2=NC(N[C@@H](CO)CC)=N1)C1=CC=CC=C1\\
    C1=CC=C(CNC2=C3C(=NC(N[C@@H](CO)CC)=N2)N(C(C)C)C=N3)C=C1\\
    CC(C)N1C=NC2=C(NCC3=CC=CC=C3)N=C(N[C@@H](CO)CC)N=C12\\
    C1=CC(CNC2=C3N=CN(C(C)C)C3=NC(N[C@H](CC)CO)=N2)=CC=C1\\
    N1(C(C)C)C2=NC(N[C@@H](CO)CC)=NC(NCC3=CC=CC=C3)=C2N=C1\\
    CC(C)N1C2=NC(N[C@@H](CO)CC)=NC(NCC3=CC=CC=C3)=C2N=C1\\
    C12=NC(N[C@H](CC)CO)=NC(NCC3=CC=CC=C3)=C1N=CN2C(C)C\\
    \lbrack C@H \rbrack(CO)(CC)NC1=NC(NCC2=CC=CC=C2)=C2C(=N1)N(C(C)C)C=N2\\
    C1=C(CNC2=C3N=CN(C(C)C)C3=NC(N[C@@H](CO)CC)=N2)C=CC=C1\\
    \end{flushleft} \\
                                                       &  & \\
                                                       \hline
                                                       & & \\
 \multirow{2}{*}{\cellcolor{color-EB9706!95}  \rotatebox[origin=c]{90} {Venetoclax~\quad}} & 
   \begin{flushleft}
  \raisebox{-0.45\height}{\includegraphics[trim={1cm 0 0 0},clip,scale=1]{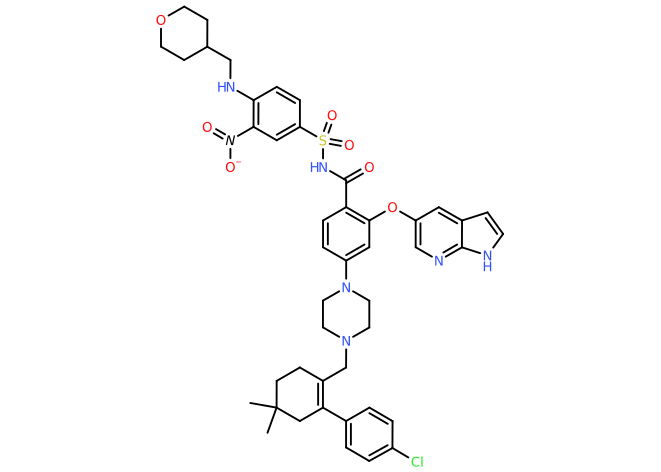}} \\ \vspace{5mm}
     CC1(C)CCC(CN2CCN(C3=CC(OC4=CN=C5C(=C4)C=CN\-5)=C(C(=O)NS(=O)(=O)C4=CC([N+](=O)[O-])=C(NCC5\-CCOCC5)C=C4)\-C=C3)CC2)=C(C2=CC=C(Cl)C=C2)C1 
    \end{flushleft}  & 
    \begin{flushleft}
C1CN(CC2=C(C3=CC=C(Cl)C=C3)CC(C)(C)CC2)CCN1C1=CC(OC2=CN=C3C(=C2)\-C=CN3)=C(C(NS(C2=CC([N+]([O-])=O)=C(NCC3CCOCC3)C=C2)(=O)=O)=O)C=C1 ~\\
C1=CC(C(=O)NS(=O)(=O)C2=CC([N+](=O)[O-])=C(NCC3CCOCC3)C=C2)=C(OC2\-=CN=C3C(=C2)C=CN3)C=C1N1CCN(CC2=C(C3=CC=C(Cl)C=C3)CC(C)(C)CC2)CC1 ~\\
\lbrack N+ \rbrack (=O)(C1=C(NCC2CCOCC2)C=CC(S(NC(=O)C2=CC=C(N3CCN(CC4=C(C5=CC=C\-(Cl)C=C5)CC(C)(C)CC4)CC3)C=C2OC2=CN=C3C(=C2)C=CN3)(=O)=O)=C1)[O-] ~\\
C1(C2=CC=C(Cl)C=C2)=C(CN2CCN(C3=CC(OC4=CN=C5NC=CC5=C4)=C(C(NS(=O)\-(C4=CC([N+]([O-])=O)=C(NCC5CCOCC5)C=C4)=O)=O)C=C3)CC2)CCC(C)(C)C1 ~\\
\lbrack O- \rbrack \lbrack N+ \rbrack (=O)C1=C(NCC2CCOCC2)C=CC(S(=O)(NC(=O)C2=CC=C(N3CCN(CC4=\-C(C5=CC=C(Cl)C=C5)CC(C)(C)CC4)CC3)C=C2OC2=CN=C3NC=CC3=C2)=O)=C1 ~\\
C1(OC2=CN=C3C(=C2)C=CN3)=CC(N2CCN(CC3=C(C4=CC=C(Cl)C=C4)CC(C)(C)\-CC3)CC2)=CC=C1C(NS(C1=CC([N+](=O)[O-])=C(NCC2CCOCC2)C=C1)(=O)=O)=O ~\\
C1(C2=C(CN3CCN(C4=CC(OC5=CN=C6C(=C5)C=CN6)=C(C(=O)NS(=O)(C5=CC([N+]\-(=O)[O-])=C(NCC6CCOCC6)C=C5)=O)C=C4)CC3)CCC(C)(C)C2)=CC=C(Cl)C=C1 ~\\
C1(N2CCN(CC3=C(C4=CC=C(Cl)C=C4)CC(C)(C)CC3)CC2)=CC(OC2=CN=C3C(=C2)C\-=CN3)=C(C(=O)NS(C2=CC([N+]([O-])=O)=C(NCC3CCOCC3)C=C2)(=O)=O)C=C1 ~\\
C1C(CNC2=C([N+](=O)[O-])C=C(S(=O)(NC(=O)C3=CC=C(N4CCN(CC5=C(C6\-=CC=C(Cl)C=C6)CC(C)(C)CC5)CC4)C=C3OC3=CN=C4NC=CC4=C3)=O)C=C2)\-CCOC1 ~\\
C12=CC(OC3=CC(N4CCN(CC5=C(C6=CC=C(Cl)C=C6)CC(C)(C)CC5)CC4)=CC=C3\-C(NS(=O)(=O)C3=CC([N+]([O-])=O)=C(NCC4CCOCC4)C=C3)=O)=CN=C1NC=C2~\\
    \end{flushleft} \\
    & \\
   \bottomrule
    \end{tabular}
    \vspace{.3cm}
    \caption{Drugs and their color highlighted for the t-SNE plots in Tables~\ref{fig:tsne-1}, \ref{fig:tsne-2} and \ref{fig:tsne-3}.}
    \label{tab:tsne-drugs}
    \vspace{-.8cm}
\end{table}

To understand the types of inductive biases introduced by the different auxiliary tasks used in pre-training, we explore the difference in behaviour of the resulting learnt representations. 





\autoref{tab:tsne-drugs} lists three well-studied compounds. We sample ten random permutations of each, along with 1000 randomly selected molecules from ChEMBL, and use this data to compare the distributions of pairwise cosine similarities of the representations learnt with each task combination in~\autoref{rdkit-ablation}.

Since the benchmark is a nearest neighbor retrieval task, we aim to understand whether learning to finely disambiguate between molecules correlates with benchmark performance. We hypothesize that models that achieve high average pairwise similarity between permutations of the same molecule and a low average pairwise similarity between random molecules achieve improved benchmark performance.


\begin{figure}[t]
    \centering
    \includegraphics[width=\textwidth]{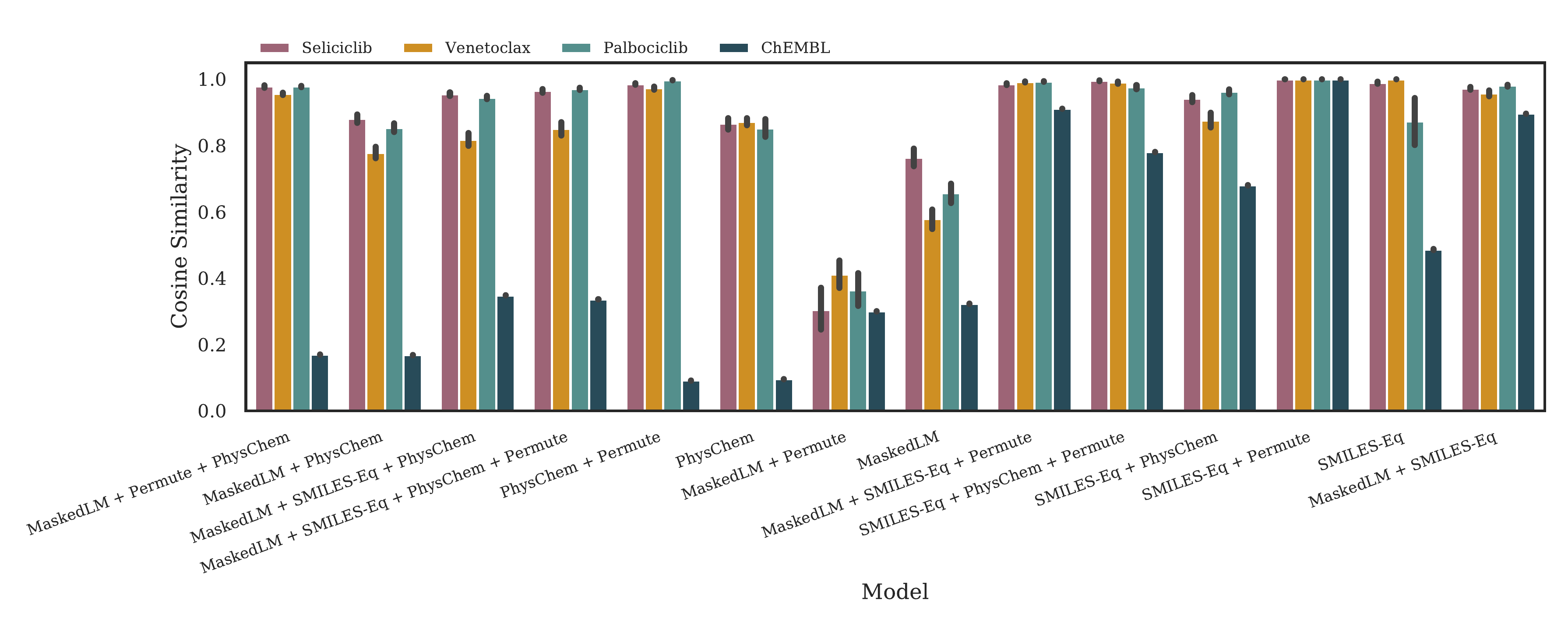}
    \caption{\label{fig:embed-sim-dist} Average pairwise cosine similarity for each group (Seliciclib, Venetoclax, Palbociclib, ChEMBL). Models are sorted from left to right by decreasing BEDROC20 as shown in~\autoref{rdkit-ablation}.}
    \label{fig:embedding_similarity}
\end{figure}

\begin{table}[h!]
    \centering
    \begin{tabular}{c c c}
    \shortstack{MaskedLM + Permute \\ + PhysChem} & PhysChem + Permute & \shortstack{MaskedLM + SMILES-Eq \\ + PhysChem + Permute} \\
    $0.323 \pm 0.0071$ & $0.293 \pm 0.071$ & $0.246 \pm 0.041$ \\
    \includegraphics[width=0.3\textwidth]{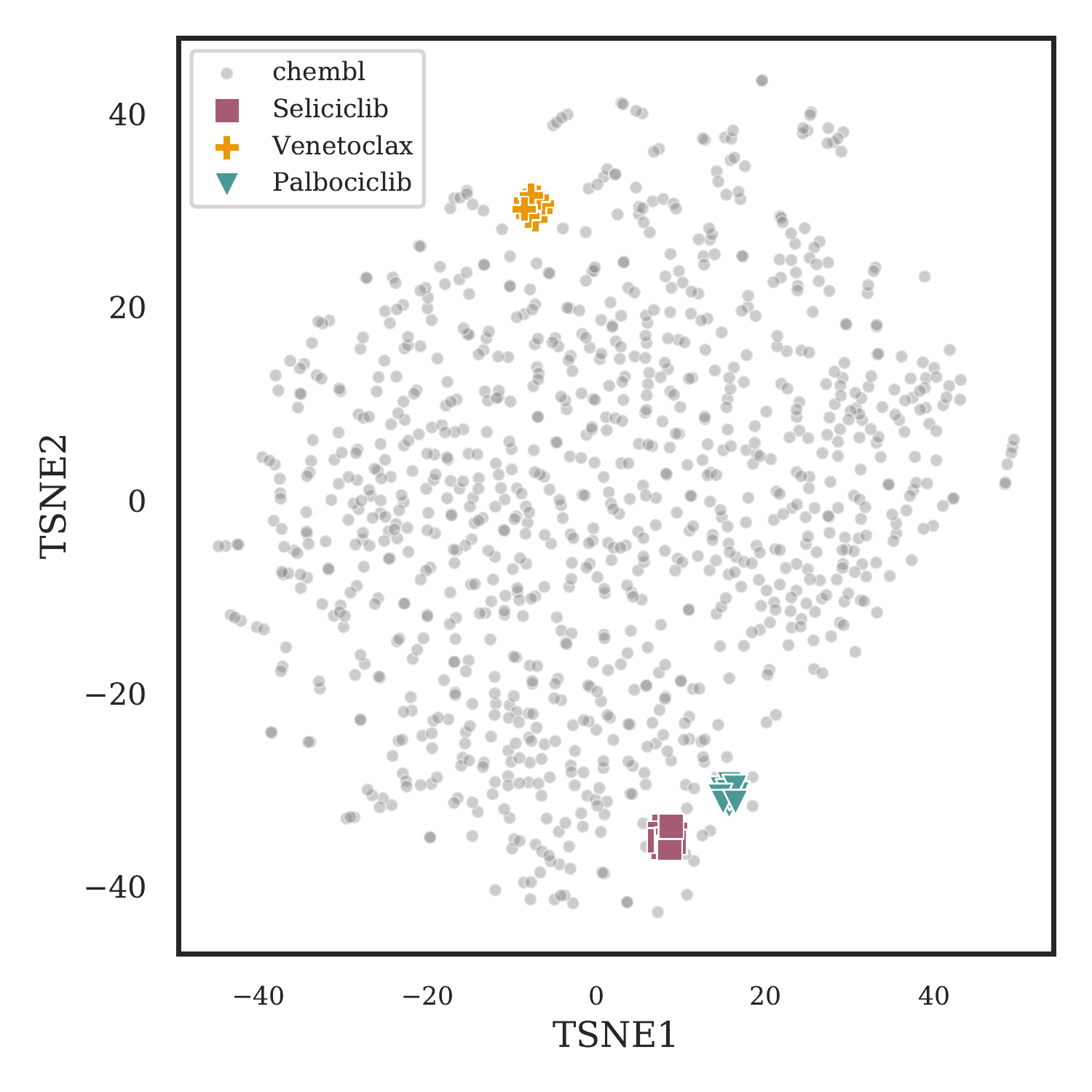} & \includegraphics[width=0.3\textwidth]{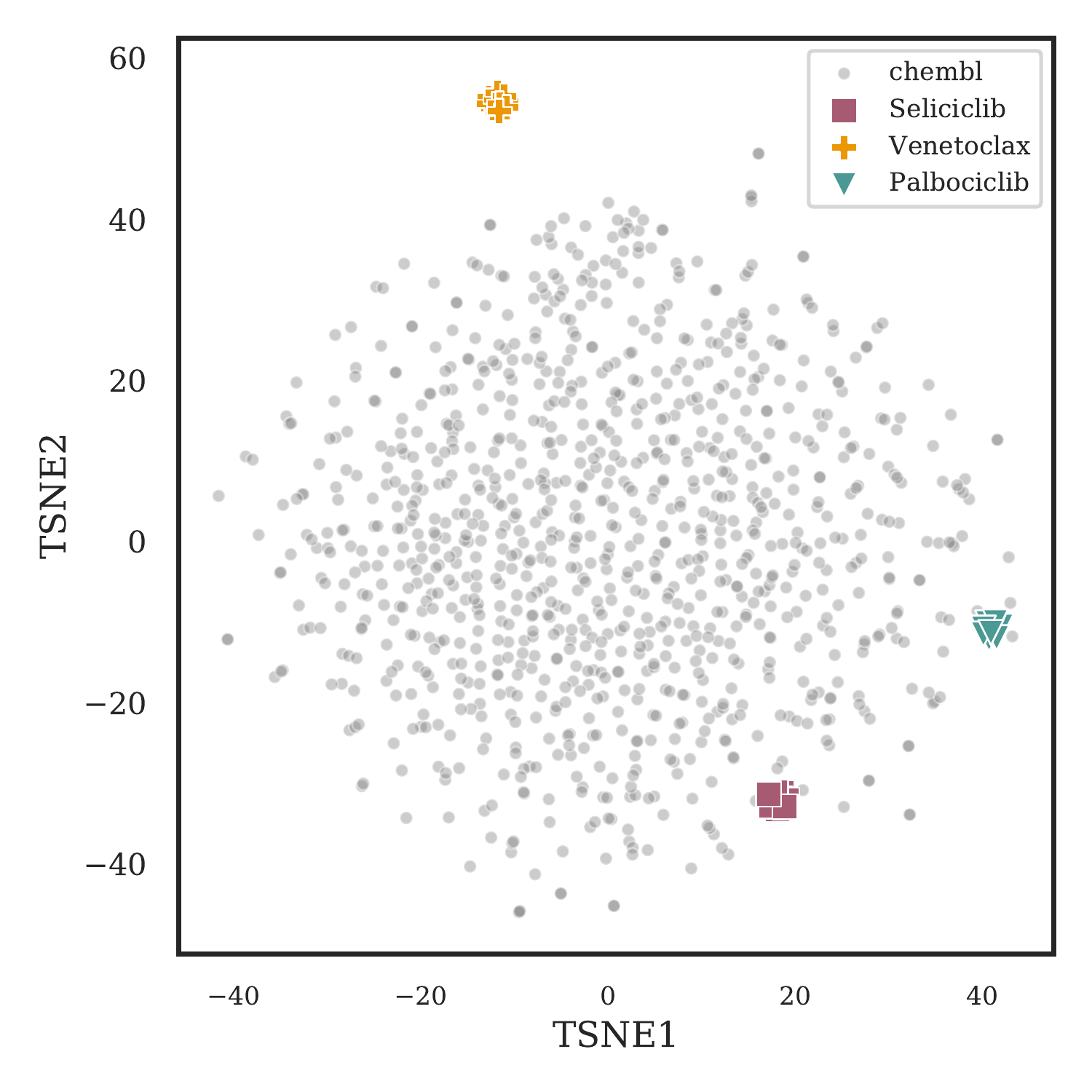} &
    \includegraphics[width=0.3\textwidth]{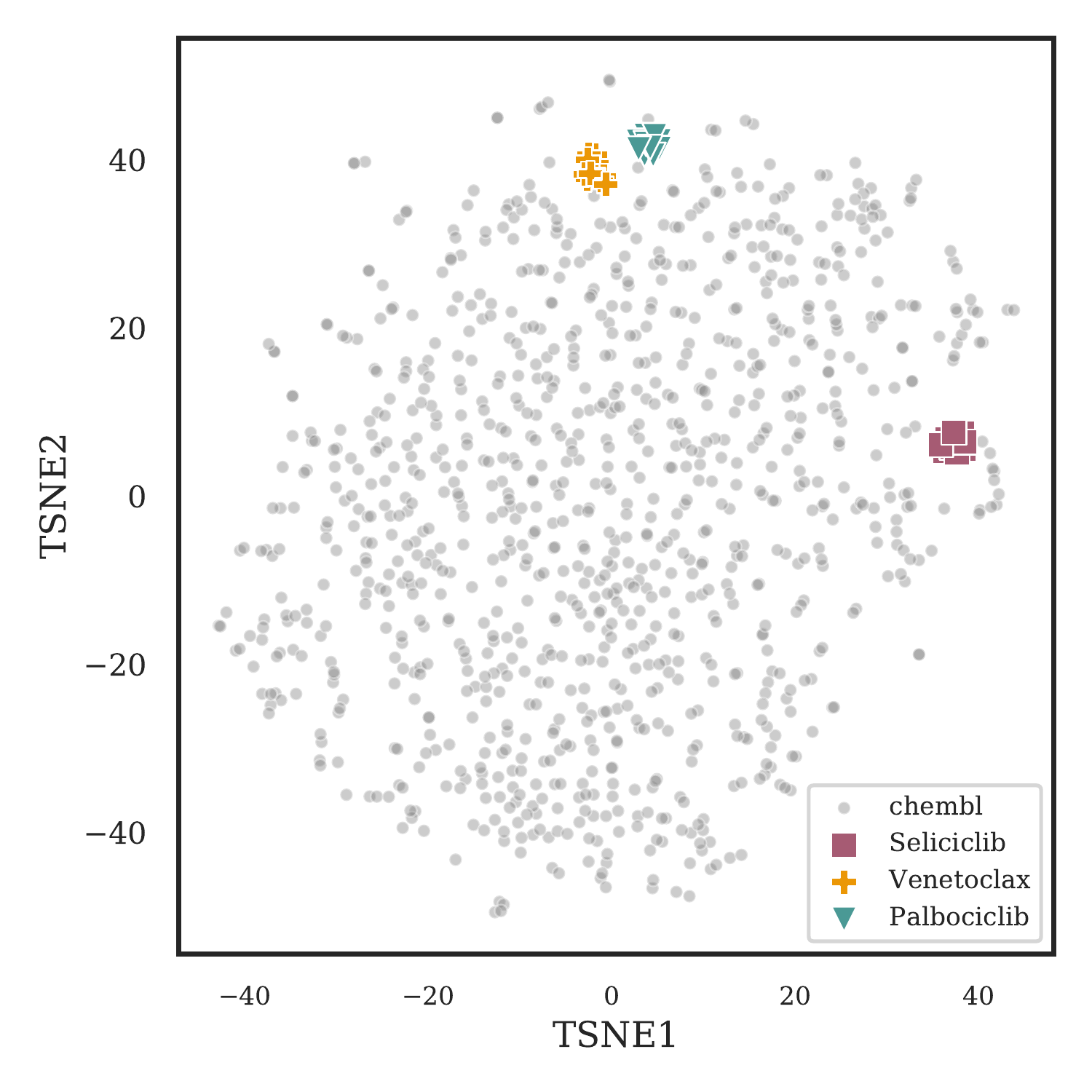}
    \end{tabular}
    \caption{t-SNE plots of molecules in~\autoref{tab:tsne-drugs}. Examples of tasks which encourage disambiguation between permutations of SMILES and others. Numbers in heading denote the BEDROC performance from~\autoref{rdkit-ablation}.}
    \label{fig:tsne-1}
\end{table}   

\begin{table}[h!]    
    \center
    \begin{tabular}{c c}
    MaskedLM   & MaskedLM + Permute   \\
    $0.250 \pm 0.073$ & $0.283 \pm 0.077$ \\
    \includegraphics[width=0.3\textwidth]{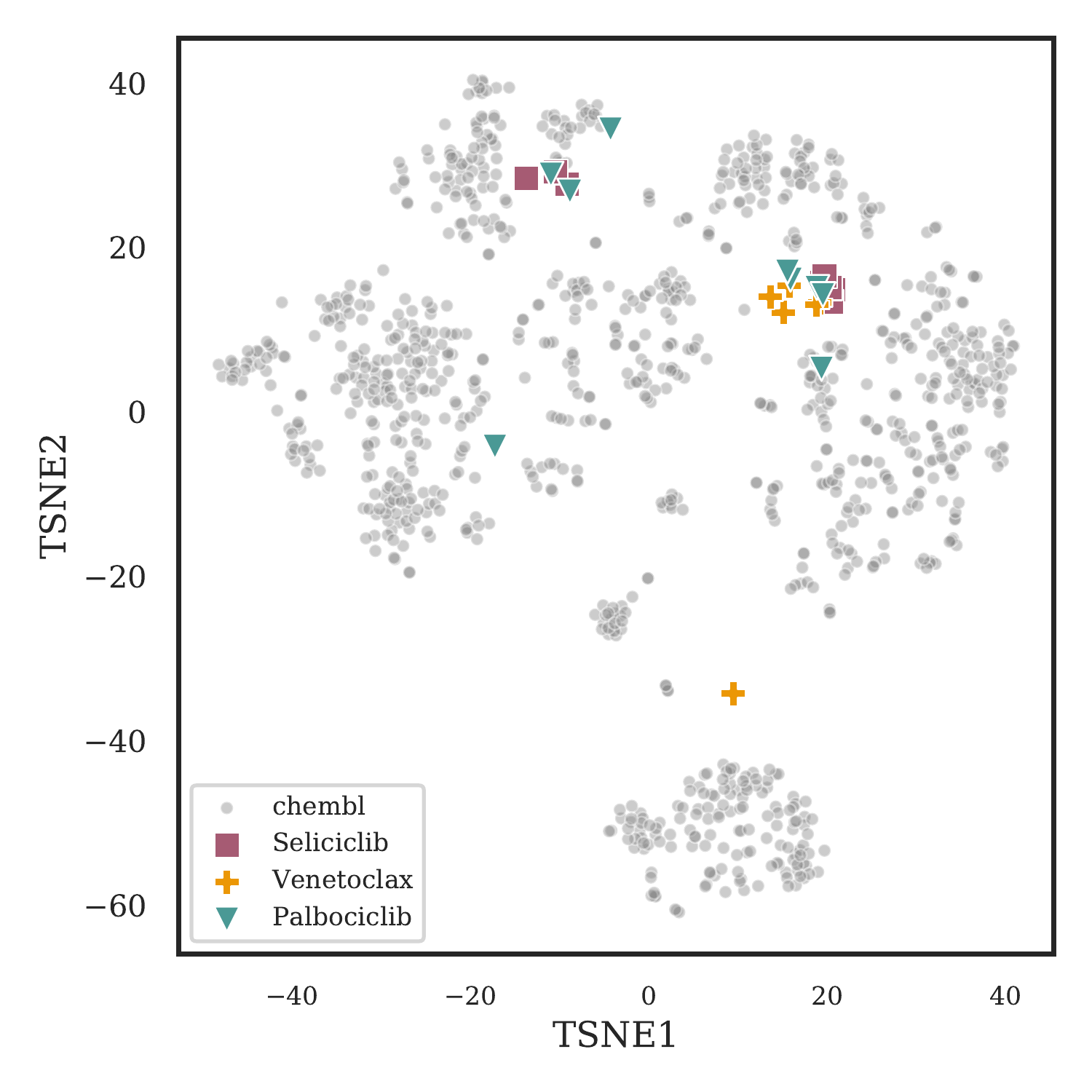} &
    \includegraphics[width=0.3\textwidth]{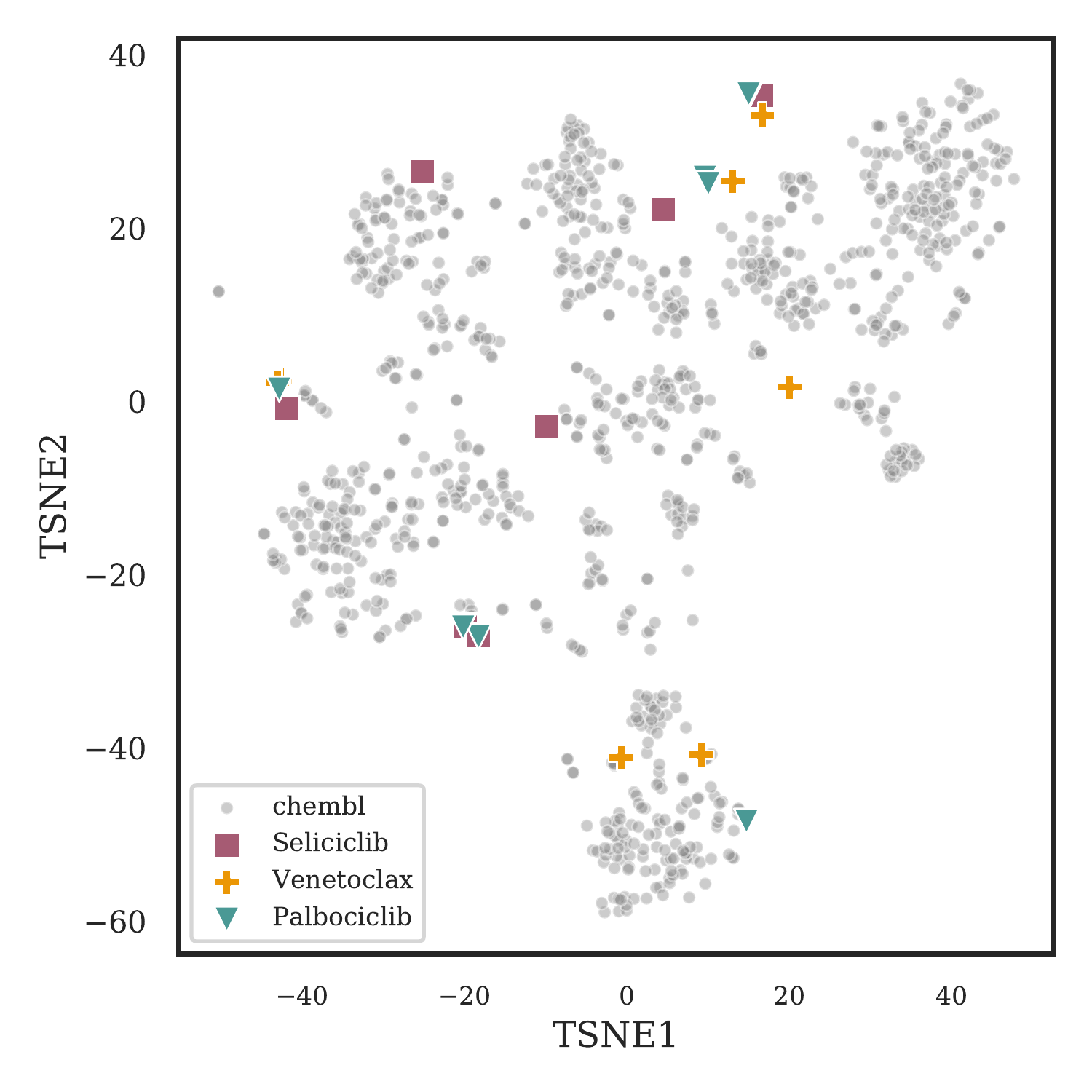} 
    \end{tabular}
    \caption{t-SNE plots of molecules in~\autoref{tab:tsne-drugs}. Examples of task combinations which don't encourage disambiguation between permutations of SMILES and others. Numbers in heading denote the BEDROC performance from~\autoref{rdkit-ablation}.}
    \label{fig:tsne-2}
 \end{table}   
 
 \begin{table}[h!]     
    \center
    \begin{tabular}{c c}
    SMILES-Eq + Permute & \shortstack{MaskedLM + SMILES-Eq \\+ Permute}  \\
    $0.005 \pm 0.037$ &  $0.092 \pm 0.069$ \\
    \includegraphics[width=0.3\textwidth]{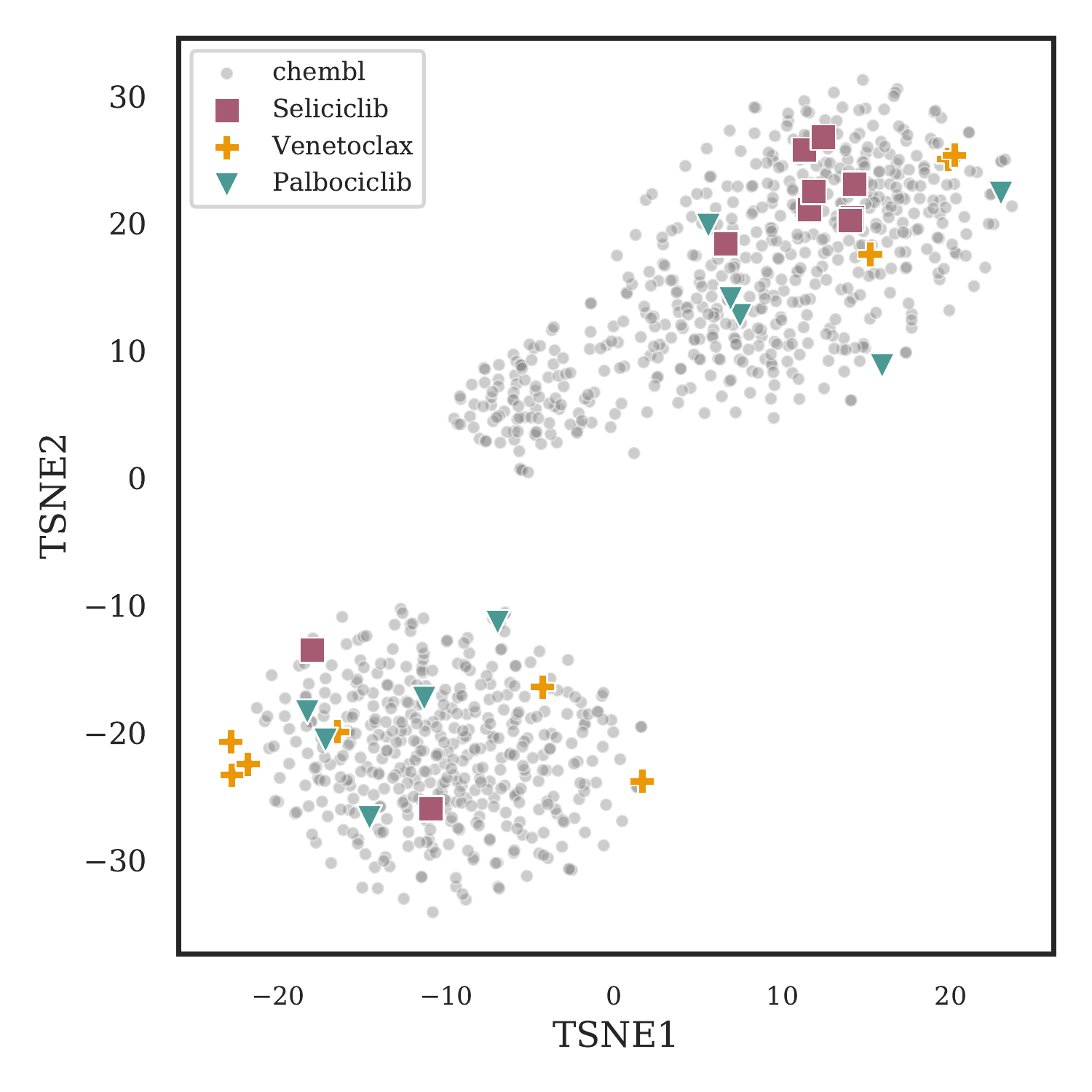} &
    \includegraphics[width=0.3\textwidth]{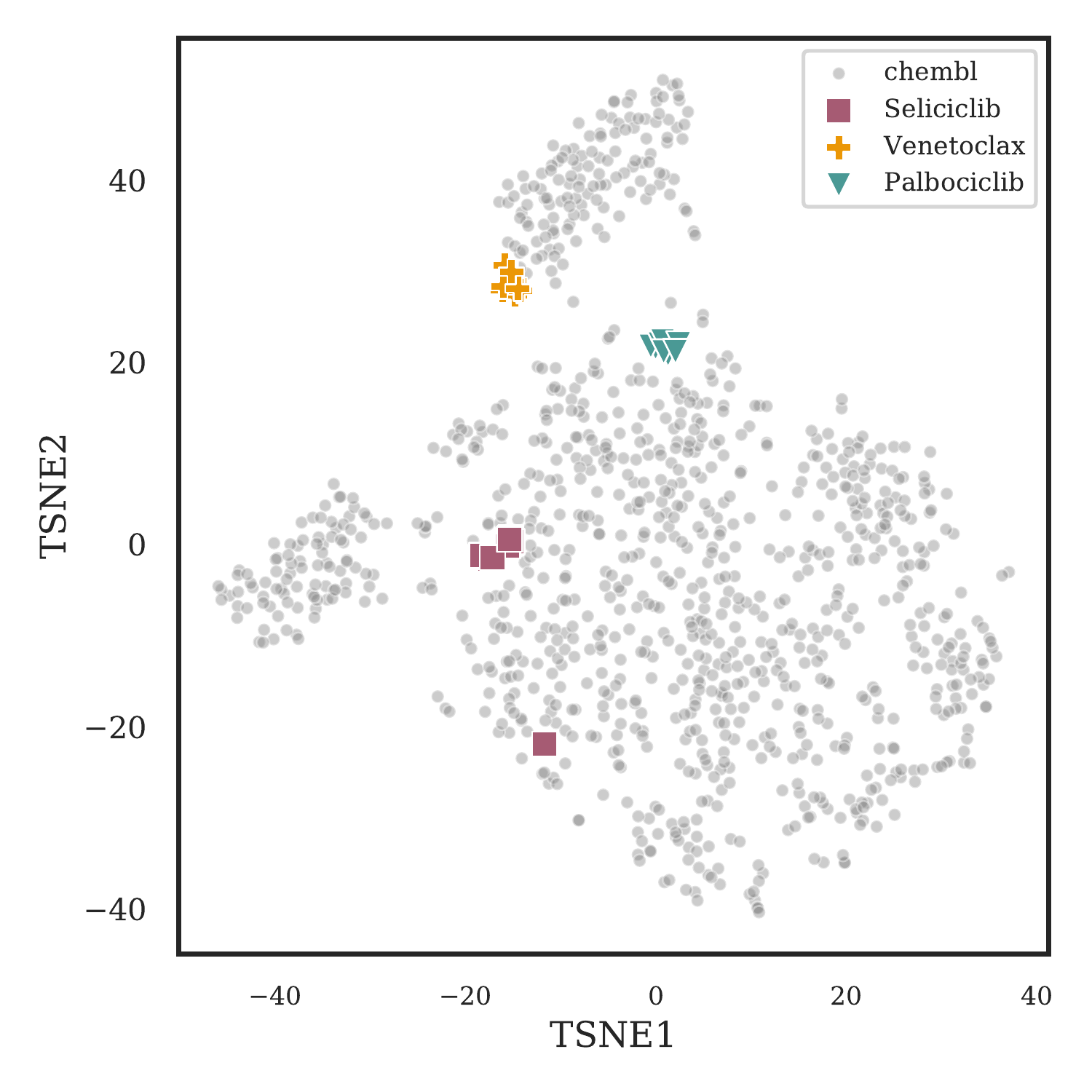} 
    \end{tabular}
     \caption{t-SNE plots of molecules in~\autoref{tab:tsne-drugs} for two models trained with the \textsc{SMILES-Eq} task. It shows that even though this task is specifically aimed at teaching a model the equivalence of permuted SMILES, it alone is not sufficient to learn a structured representation space. Numbers in heading denote the BEDROC performance from~\autoref{rdkit-ablation}.}
    \label{fig:tsne-3}
\end{table}

\autoref{fig:embedding_similarity} gives the results of our analysis where we observe three broad behaviours: First, \molbert~models that were pre-trained on task combinations including the physicochemical property prediction (\textsc{PhysChemPred}) task successfully assign high similarities to permutations of the same molecule, while assigning low average similarity between randomly selected ChEMBL compounds. This suggests that tasks that encourage this large margin property help models organize the embedding space in a more semantically meaningful manner.

The \molbert~representations resulting from a sample of tasks with this characteristic are given in~\autoref{fig:tsne-1}. For each task we plot t-SNE \citep{maaten2008visualizing} projections of \molbert~representations for the 1000 randomly selected molecules from ChEMBL in gray and each of the three molecules listed in~\autoref{tab:tsne-drugs} along with their ten SMILES permutations. We use the \texttt{sklearn} implementation of t-SNE~\citep{scikit-learn} with parameters: $perplexity=30, early\_exaggeration = 12, learning\_rate = 200$.


Second, two task combinations \textsc{MaskedLM} and \textsc{MaskedLM + Permute} lead to models that assign very low similarities to both the permutations of the same molecule, and to other molecules. 
This suggests that these combinations encourage models to map inputs to a large representation space, but fail to structure this space in a semantically meaningful way -- observed in the two sets of similarities being almost equal. See~\autoref{fig:tsne-2}.

Finally, explicit encouragement of the model to learn to recognise permutations of identical SMILES in the form of the \textsc{Smiles-Eq} task does not seem to be sufficient to enable models to increase the average margin between representations of identical permutations and unrelated molecules. \autoref{fig:tsne-3} shows that the \textsc{SMILES-Eq + Permute} combination is not able to generate a structured representation space. However, when the \textsc{MaskedLM} task is added, the embeddings of the different permutations are grouped together. This also results in an improved BEDROC from~\autoref{rdkit-ablation}.

\section{PhysChemPred subset ablation}\label{app:physchem}
\begin{table}[t]
    \centering
    \begin{tabular}{cc|cc}
        \toprule
         &  $n$    &    AUROC &       BEDROC20 \\
        \midrule
                   ALL & 200 & $\mathbf{0.738 \pm 0.060}$ &  $\mathbf{0.323 \pm 0.071}$ \\
               SURFACE & 49 & $\mathbf{0.738 \pm 0.061}$ &  $0.310 \pm 0.056$ \\
                CHARGE & 18 & $0.711 \pm 0.063$ &  $0.270 \pm 0.074$ \\
              FRAGMENT & 101 & $0.704 \pm 0.065$ &  $0.277 \pm 0.070$ \\
                SIMPLE & 8 & $0.695 \pm 0.065$ &  $0.255 \pm 0.060$ \\
                 GRAPH & 19 & $0.693 \pm 0.066$ &  $0.249 \pm 0.067$ \\
                ESTATE & 25 & $0.676 \pm 0.063$ &  $0.232 \pm 0.070$ \\
          DRUGLIKENESS & 24 & $0.671 \pm 0.060$ &  $0.214 \pm 0.069$ \\
                  LOGP & 13 & $0.651 \pm 0.064$ &  $0.186 \pm 0.064$ \\
          REFRACTIVITY & 11 & $0.649 \pm 0.063$ &  $0.193 \pm 0.058$ \\
               GENERAL & 12 & $0.633 \pm 0.064$ &  $0.201 \pm 0.069$ \\
        \bottomrule
    \end{tabular}
    \caption{AUROC and BEDROC20 on the RDKit virtual screening benchmark for \molbert~trained with various subsets of the RDKit calculated physicochemical properties.}
    \label{subset-ablation}
\end{table}

\autoref{subset-ablation} gives an ablation of the \textsc{PhysChemPred} task by grouping the $200$ descriptors used into smaller subsets of related descriptors, and repeating the Virtual Screening benchmark.

From the table we observe that in general all physicochemical descriptors do well on the benchmark. Moreover, some subsets which contain very few descriptors (\textsc{SURFACE} $n=49$ and \textsc{CHARGE} $n=18$ ) are able to achieve almost the same results as using all of physicochemical descriptors.

Finally, the lists of descriptors and their groupings are as follows:

\begin{description}
    [leftmargin=0.3cm,
    style=unboxed,
    labelsep=-0.05cm,
    itemsep=-0.05cm,
    align=left,  font=\bf]
    
    \item [ALL:~] ~Full set of 200 descriptors from RDKit. \\
    {\small \texttt{BalabanJ, BertzCT, Chi0, Chi0n, Chi0v, Chi1, Chi1n, Chi1v, Chi2n, Chi2v, Chi3n, Chi3v, Chi4n, Chi4v, EState\_VSA1, EState\_VSA10, EState\_VSA11, EState\_VSA2, EState\_VSA3, EState\_VSA4, EState\_VSA5, EState\_VSA6, EState\_VSA7, EState\_VSA8, EState\_VSA9, ExactMolWt, FpDensityMorgan1, FpDensityMorgan2, FpDensityMorgan3, FractionCSP3, HallKierAlpha, HeavyAtomCount, HeavyAtomMolWt, Ipc, Kappa1, Kappa2, Kappa3, LabuteASA, MaxAbsEStateIndex, MaxAbsPartialCharge, MaxEStateIndex, MaxPartialCharge, MinAbsEStateIndex, MinAbsPartialCharge, MinEStateIndex, MinPartialCharge, MolLogP, MolMR, MolWt, NHOHCount, NOCount, NumAliphaticCarbocycles, NumAliphaticHeterocycles, NumAliphaticRings, NumAromaticCarbocycles, NumAromaticHeterocycles, NumAromaticRings, NumHAcceptors, NumHDonors, NumHeteroatoms, NumRadicalElectrons, NumRotatableBonds, NumSaturatedCarbocycles, NumSaturatedHeterocycles, NumSaturatedRings, NumValenceElectrons, PEOE\_VSA1, PEOE\_VSA10, PEOE\_VSA11, PEOE\_VSA12, PEOE\_VSA13, PEOE\_VSA14, PEOE\_VSA2, PEOE\_VSA3, PEOE\_VSA4, PEOE\_VSA5, PEOE\_VSA6, PEOE\_VSA7, PEOE\_VSA8, PEOE\_VSA9, RingCount, SMR\_VSA1, SMR\_VSA10, SMR\_VSA2, SMR\_VSA3, SMR\_VSA4, SMR\_VSA5, SMR\_VSA6, SMR\_VSA7, SMR\_VSA8, SMR\_VSA9, SlogP\_VSA1, SlogP\_VSA10, SlogP\_VSA11, SlogP\_VSA12, SlogP\_VSA2, SlogP\_VSA3, SlogP\_VSA4, SlogP\_VSA5, SlogP\_VSA6, SlogP\_VSA7, SlogP\_VSA8, SlogP\_VSA9, TPSA, VSA\_EState1, VSA\_EState10, VSA\_EState2, VSA\_EState3, VSA\_EState4, VSA\_EState5, VSA\_EState6, VSA\_EState7, VSA\_EState8, VSA\_EState9, fr\_Al\_COO, fr\_Al\_OH, fr\_Al\_OH\_noTert, fr\_ArN, fr\_Ar\_COO, fr\_Ar\_N, fr\_Ar\_NH, fr\_Ar\_OH, fr\_COO, fr\_COO2, fr\_C\_O, fr\_C\_O\_noCOO, fr\_C\_S, fr\_HOCCN, fr\_Imine, fr\_NH0, fr\_NH1, fr\_NH2, fr\_N\_O, fr\_Ndealkylation1, fr\_Ndealkylation2, fr\_Nhpyrrole, fr\_SH, fr\_aldehyde, fr\_alkyl\_carbamate, fr\_alkyl\_halide, fr\_allylic\_oxid, fr\_amide, fr\_amidine, fr\_aniline, fr\_aryl\_methyl, fr\_azide, fr\_azo, fr\_barbitur, fr\_benzene, fr\_benzodiazepine, fr\_bicyclic, fr\_diazo, fr\_dihydropyridine, fr\_epoxide, fr\_ester, fr\_ether, fr\_furan, fr\_guanido, fr\_halogen, fr\_hdrzine, fr\_hdrzone, fr\_imidazole, fr\_imide, fr\_isocyan, fr\_isothiocyan, fr\_ketone, fr\_ketone\_Topliss, fr\_lactam, fr\_lactone, fr\_methoxy, fr\_morpholine, fr\_nitrile, fr\_nitro, fr\_nitro\_arom, fr\_nitro\_arom\_nonortho, fr\_nitroso, fr\_oxazole, fr\_oxime, fr\_para\_hydroxylation, fr\_phenol, fr\_phenol\_noOrthoHbond, fr\_phos\_acid, fr\_phos\_ester, fr\_piperdine, fr\_piperzine, fr\_priamide, fr\_prisulfonamd, fr\_pyridine, fr\_quatN, fr\_sulfide, fr\_sulfonamd, fr\_sulfone, fr\_term\_acetylene, fr\_tetrazole, fr\_thiazole, fr\_thiocyan, fr\_thiophene, fr\_unbrch\_alkane, fr\_urea, qed}}
    \vspace{1mm}
    
    \item [SURFACE:~] ~MOE-based surface descriptor subset. \\
    {\small \texttt{EState\_VSA1, EState\_VSA10, EState\_VSA11, EState\_VSA2, EState\_VSA3, EState\_VSA4, EState\_VSA5, EState\_VSA6, EState\_VSA7, EState\_VSA8, EState\_VSA9, LabuteASA, PEOE\_VSA1, PEOE\_VSA10, PEOE\_VSA11, PEOE\_VSA12, PEOE\_VSA13, PEOE\_VSA14, PEOE\_VSA2, PEOE\_VSA3, PEOE\_VSA4, PEOE\_VSA5, PEOE\_VSA6, PEOE\_VSA7, PEOE\_VSA8, PEOE\_VSA9, SMR\_VSA1, SMR\_VSA10, SMR\_VSA2, SMR\_VSA3, SMR\_VSA4, SMR\_VSA5, SMR\_VSA6, SMR\_VSA7, SMR\_VSA8, SMR\_VSA9, SlogP\_VSA1, SlogP\_VSA10, SlogP\_VSA11, SlogP\_VSA12, SlogP\_VSA2, SlogP\_VSA3, SlogP\_VSA4, SlogP\_VSA5, SlogP\_VSA6, SlogP\_VSA7, SlogP\_VSA8, SlogP\_VSA9, TPSA}}
    \vspace{1mm}
    
    \item [CHARGE:~] ~Partial charge and VSA/charge descriptor subset. \\
    {\small \texttt{MaxAbsPartialCharge, MaxPartialCharge, MinAbsPartialCharge, MinPartialCharge, PEOE\_VSA1, PEOE\_VSA10, PEOE\_VSA11, PEOE\_VSA12, PEOE\_VSA13, PEOE\_VSA14, PEOE\_VSA2, PEOE\_VSA3, PEOE\_VSA4, PEOE\_VSA5, PEOE\_VSA6, PEOE\_VSA7, PEOE\_VSA8, PEOE\_VSA9}}
    \vspace{1mm}
    
    \item [FRAGMENT:~] ~Count and fragment based descriptor subset. \\
    {\small \texttt{NHOHCount, NOCount, NumAliphaticCarbocycles, NumAliphaticHeterocycles, NumAliphaticRings, NumAromaticCarbocycles, NumAromaticHeterocycles, NumAromaticRings, NumHAcceptors, NumHDonors, NumHeteroatoms, NumRotatableBonds, NumSaturatedCarbocycles, NumSaturatedHeterocycles, NumSaturatedRings, RingCount, fr\_Al\_COO, fr\_Al\_OH, fr\_Al\_OH\_noTert, fr\_ArN, fr\_Ar\_COO, fr\_Ar\_N, fr\_Ar\_NH, fr\_Ar\_OH, fr\_COO, fr\_COO2, fr\_C\_O, fr\_C\_O\_noCOO, fr\_C\_S, fr\_HOCCN, fr\_Imine, fr\_NH0, fr\_NH1, fr\_NH2, fr\_N\_O, fr\_Ndealkylation1, fr\_Ndealkylation2, fr\_Nhpyrrole, fr\_SH, fr\_aldehyde, fr\_alkyl\_carbamate, fr\_alkyl\_halide, fr\_allylic\_oxid, fr\_amide, fr\_amidine, fr\_aniline, fr\_aryl\_methyl, fr\_azide, fr\_azo, fr\_barbitur, fr\_benzene, fr\_benzodiazepine, fr\_bicyclic, fr\_diazo, fr\_dihydropyridine, fr\_epoxide, fr\_ester, fr\_ether, fr\_furan, fr\_guanido, fr\_halogen, fr\_hdrzine, fr\_hdrzone, fr\_imidazole, fr\_imide, fr\_isocyan, fr\_isothiocyan, fr\_ketone, fr\_ketone\_Topliss, fr\_lactam, fr\_lactone, fr\_methoxy, fr\_morpholine, fr\_nitrile, fr\_nitro, fr\_nitro\_arom, fr\_nitro\_arom\_nonortho, fr\_nitroso, fr\_oxazole, fr\_oxime, fr\_para\_hydroxylation, fr\_phenol, fr\_phenol\_noOrthoHbond, fr\_phos\_acid, fr\_phos\_ester, fr\_piperdine, fr\_piperzine, fr\_priamide, fr\_prisulfonamd, fr\_pyridine, fr\_quatN, fr\_sulfide, fr\_sulfonamd, fr\_sulfone, fr\_term\_acetylene, fr\_tetrazole, fr\_thiazole, fr\_thiocyan, fr\_thiophene, fr\_unbrch\_alkane, fr\_urea}}
    \vspace{1mm}
    
    \item [SIMPLE:~] Small set of commonly used descriptors. \\
    {\small \texttt{FpDensityMorgan2, FractionCSP3, MolLogP, MolWt, NumHAcceptors, NumHDonors, NumRotatableBonds, TPSA} }
    \vspace{1mm}
    
    \item [GRAPH:~] ~Graph descriptor subset (following the grouping found in the $GraphDescriptors$ module in RDKit). \\
    {\small \texttt{BalabanJ, BertzCT, Chi0, Chi0n, Chi0v, Chi1, Chi1n, Chi1v, Chi2n, Chi2v, Chi3n, Chi3v, Chi4n, Chi4v, HallKierAlpha, Ipc, Kappa1, Kappa2, Kappa3}}
    \vspace{1mm}
    
    \item [ESTATE:~] ~Electrotopological state (e-state) and VSA/e-state descriptor subset. \\
    {\small \texttt{EState\_VSA1, EState\_VSA10, EState\_VSA11, EState\_VSA2, EState\_VSA3, EState\_VSA4, EState\_VSA5, EState\_VSA6, EState\_VSA7, EState\_VSA8, EState\_VSA9, MaxAbsEStateIndex, MaxEStateIndex, MinAbsEStateIndex, MinEStateIndex, VSA\_EState1, VSA\_EState10, VSA\_EState2, VSA\_EState3, VSA\_EState4, VSA\_EState5, VSA\_EState6, VSA\_EState7, VSA\_EState8, VSA\_EState9}}
    \vspace{1mm}
    
    \item [DRUGLIKENESS:~] ~Subset of descriptors commonly used to assess druglikeness. \\
    {\small \texttt{ExactMolWt, FractionCSP3, HeavyAtomCount, MolLogP, MolMR, MolWt, NHOHCount, NOCount, NumAliphaticCarbocycles, NumAliphaticHeterocycles, NumAliphaticRings, NumAromaticCarbocycles, NumAromaticHeterocycles, NumAromaticRings, NumHAcceptors, NumHDonors, NumHeteroatoms, NumRotatableBonds, NumSaturatedCarbocycles, NumSaturatedHeterocycles, NumSaturatedRings, RingCount, TPSA, qed}}
    \vspace{1mm}
    
    \item [LOGP:~] ~LogP and VSA/LogP descriptor subset. \\
    {\small \texttt{MolLogP, SlogP\_VSA1, SlogP\_VSA10, SlogP\_VSA11, SlogP\_VSA12, SlogP\_VSA2, SlogP\_VSA3, SlogP\_VSA4, SlogP\_VSA5, SlogP\_VSA6, SlogP\_VSA7, SlogP\_VSA8, SlogP\_VSA9}}
    \vspace{1mm}
    
    \item [REFRACTIVITY:~] ~MOE-based refractivity related descriptor subset. \\
    {\small \texttt{MolMR, SMR\_VSA1, SMR\_VSA10, SMR\_VSA2, SMR\_VSA3, SMR\_VSA4, SMR\_VSA5, SMR\_VSA6, SMR\_VSA7, SMR\_VSA8, SMR\_VSA9}}
    \vspace{1mm}
    
    \item [GENERAL:~] ~General descriptor subset (following the grouping found in the $Descriptors$ module in RDKit).\\
    {\small \texttt{ExactMolWt, FpDensityMorgan1, FpDensityMorgan2, FpDensityMorgan3, HeavyAtomMolWt, MaxAbsPartialCharge, MaxPartialCharge, MinAbsPartialCharge, MinPartialCharge, MolWt, NumRadicalElectrons, NumValenceElectrons}}
\end{description}

\end{document}